\documentclass[a4paper, fleqn]{cas-sc}

\usepackage[numbers]{natbib}

\usepackage{booktabs}
\usepackage{multirow}
\usepackage{float}
\usepackage{mathtools}
\usepackage{soul}
\usepackage{color}
\usepackage{hyperref}
\usepackage{diagbox}
\usepackage{makecell}
\usepackage[toc,page]{appendix}
\usepackage[colorinlistoftodos]{todonotes}
\usepackage{subfigure}
\usepackage{amsmath}
\usepackage{amssymb}
\usepackage{mathrsfs}
\usepackage{bm}
\usepackage{lineno}
\usepackage{subfigure}
\usepackage[ruled,linesnumbered]{algorithm2e}
\usepackage{caption}
\begin{document}

\let\WriteBookmarks\relax
\def\floatpagepagefraction{1}
\def\textpagefraction{.001}
\shorttitle{MTS Classification with Graph Pooling}
\shortauthors{Ziheng Duan and Haoyan Xu et~al.}

\title [mode = title]{Multivariate Time Series Classification with Hierarchical Variational Graph Pooling}       

\address[1]{School of Big Data and Software Engineering, Chongqing University, Chongqing, 401331, China}
\address[2]{College of Energy Engineering, Zhejiang University, Zhejiang, 310027, China}
\address[3]{Department of Computer Science, University of California, Los Angeles, CA 90095, USA}

\cortext[cor1]{These authors contributed equally to this research.}
\cortext[cor2]{Corresponding author.}

\author[1,2]{Ziheng Duan}
\cormark[1]

\author[1,2,3]{Haoyan Xu}
\cormark[1]

\author[1]{Yueyang Wang}[orcid=0000-0003-3210-0930]
\cormark[2]

\author[1]{Yida Huang}

\author[2]{Anni Ren}

\author[2]{Zhongbin Xu}

\author[3]{Yizhou Sun}

\author[3]{Wei Wang}

\begin{abstract}
With the advancement of sensing technology, multivariate time series classification (MTSC) has recently received considerable attention.
Existing deep learning-based MTSC techniques, which mostly rely on convolutional or recurrent neural networks, are primarily concerned with the temporal dependency of single time series. As a result, they struggle to express pairwise dependencies among multivariate variables directly.
Furthermore, current spatial-temporal modeling (e.g., graph classification) methodologies based on Graph Neural Networks (GNNs) are inherently flat and cannot aggregate hub data in a hierarchical manner.
To address these limitations, we propose a novel graph pooling-based framework MTPool to obtain the expressive global representation of MTS.
We first convert MTS slices to graphs by utilizing interactions of variables via graph structure learning module and attain the spatial-temporal graph node features via temporal convolutional module.
To get global graph-level representation, we design an "encoder-decoder" based variational graph pooling module for creating adaptive centroids for cluster assignments.
Then we combine GNNs and our proposed variational graph pooling layers for joint graph representation learning and graph coarsening, after which the graph is progressively coarsened to one node. 
At last, a differentiable classifier takes this coarsened representation to get the final predicted class. 
Experiments on ten benchmark datasets exhibit MTPool outperforms state-of-the-art strategies in the MTSC task.

\end{abstract}


\maketitle

\section{Introduction}

Multivariate time series (MTS), which are gathered from numerous variables or sensors in our day-by-day life, are utilized in different investigations \cite{li2021dtdr,shi2021parallel,feng2021hybrid,zhou2020parallel}.
Attributable to cutting edge sensing techniques, the Multivariate Time Series Classification (MTSC) issue, recognizing the labels for MTS records, has pulled in a ton of consideration in late numerous years \cite{zhang2020tapnet,baldan2019distributed,mori2019early}. 
MTSC models have been applied in a broad range of real-world applications \cite{,zhou2021comparative,iwana2020dtw}, such as sleep stage identification \cite{sun2019hierarchical}, healthcare \cite{kang2014bayesian} and action recognition \cite{yu2015real}.
Specifically, MTS has the accompanying two significant attributes: (1) Each univariate time series has an \emph{inner temporal reliance} mode; (2) There always exist \emph{hidden dependency} relationships among different MTS variables. Capturing these two attributes is a significant contribution to getting better classification performance but also a challenging task.

A lot of MTS classification methods have been proposed throughout the long term. 
Distance-based methods, like Dynamic Time Warping (DTW) with k-NN \cite{seto2015multivariate}, and feature-based methods such as Hidden Unit Logistic Model (HULM) \cite{pei2017multivariate} have proven to be successful in classification tasks on many benchmark MTS datasets.
In any case, these methodologies need hefty crafting on data preprocessing and feature engineering and they cannot fully explore the \emph{inner temporal reliance} mode in each univariate time series.
Recently, many deep learning-based methods have been exploited for end-to-end MTS classification.
Fully convolutional networks (FCN) and the residual networks (ResNet) can achieve comparable or better performance than traditional methods \cite{wang2017time}. 
MLSTM-FCN \cite{karim2019multivariate} utilizes an LSTM layer and a stacked CNN layer alongside squeeze-and-excitation blocks to obtain representations.
These deep learning-based methods have achieved promising performance in MTSC tasks. 
Notwithstanding, the current deep learning-based strategies don't show the \emph{hidden dependency} relationships among different MTS variables. Their input of these models should be grid data, also limiting the representational ability.

To address the above problem, we first construct graphs from MTS instead of directly utilizing the sequences of MTS.
Thus, variables from MTS are constructed as nodes in the converted graph, and they are interlinked through their hidden relations. 
As we know, graphs are a particular type of information that portrays the relationships between various entities or nodes.
Existing mining graph methods, like Graph Neural Networks (GNNs) \cite{scarselli2008graph}, aggregating feature information of adjoining nodes to learn node or graph representation, have been demonstrated successfully capturing the hidden relations of nodes.
Consequently, mining MTS information by utilizing graph neural networks can be a promising method to save their temporal patterns while exploiting the interdependency among time series variables \cite{xu2020multivariate,wang2020mthetgnn,wu2020connecting}.
Therefore, this paper proposes converting MTS slices to graphs through graph structure learning and viewing the MTS classification task as a graph classification task.

The graph classification task aims to predict the type of the whole graph, where the graph structure and all the initial node-level representations are inputs.
For instance, given a molecule, the task could be to anticipate whether it is poisonous \cite{ranjan2020asap}. 
Although current GNNs based spatial-temporal modeling strategies could aggregate the graph structure and node-level representations, they are still inherently flat and do not have the capacity of aggregate hub data in a hierarchical way \cite{ying2018hierarchical}.
Recently, some hierarchical pooling methods have been proposed and achieved promising results in many graph classification tasks, such as gPool [7], DiffPool [34], MemGNN [12]. These methods design specific hierarchical graph pooling layers, where nodes are recursively aggregated to frame a cluster addressing a hub in the pooled graph. 
gPool downsamples by selecting the most important nodes. DiffPool downsamples by clustering the nodes using GNNs.
Like DiffPool, MemGNN utilizes a multi-head exhibit of memory keys and a convolution operator to sum the soft cluster assignments from various heads and calculate the attention scores among nodes and clusters.
However, we notice that the key process of most hierarchical pooling mechanisms, generating centroids for soft cluster assignments, is not related to input graphs. This does not make sense because the centroid for different graphs should be different. Although MemGNN randomly initializes centroids (keys) before training and makes centroids learnable during training. The centroids cannot change when testing so that the testing graphs can only use the same centroids as train graphs for cluster assignments. 
Intuitively, each input graph should have its specific centroids based on its topology structure and node features. 
In general, the pooling mechanism should keep three significant properties of graphs: 
(1) \emph{Permutation invariance}: The centroids of the same graph should be invariant when the permutation of nodes changes.
(2) \emph{Input correlation}: The centroids should change if the input graph changes.
(3) \emph{Dimensional adjustability}. The dimension of the centroids matrix should be adjustable by the dimension of the input graph and coarsened graph.

Towards this end, we propose a novel MTS classification framework called MTPool (\emph{\underline{M}}ultivariate \emph{\underline{T}}ime Series Classification with Variational Graph \emph{\underline{Pool}}ing).
The goal of MTPool is to model temporal patterns and hidden dependency relationships among MTS variables and obtain the MTS's global representation. Specifically, MTPool first constructs a graph based on MTS data through the graph structure learning module and learns spatial-temporal features of MTS through the temporal convolution module. Then, to keep the three properties mentioned above, we design a novel pooling layer, Variational Pooling, for graph coarsening. This pooling layer contains an "encoder-decoder" architecture, enabling the generation process of centroids to be input-related and making the model more inductive. Next, MTPool adapts GNNs for jointly learning graph representation, after which the graph is progressively coarsened to one node. Finally, these final coarsened graph-level representations can be used as features input to a differentiable classifier for MTSC tasks.
To summarize, our fundamental contributions can be finished up as follows:

\begin{enumerate}[(1)]
\item As far as we could possibly know, we first propose a hierarchical graph pooling-based framework to model MTS and hierarchically generate its global representation for MTS classification.
\item We design MTPool as an end-to-end joint framework for graph structure learning, temporal convolution, graph representation learning, and graph coarsening.
\item We propose a novel pooling method, Variational Pooling. The centroids for cluster assignments are input-related to the input graphs, making the model more inductive and leading to better performance.
\item We conduct extensive experiments on MTS benchmark datasets. Experiments on ten benchmark datasets exhibit MTPool outperforms cutting edge strategies in the MTSC task.
\end{enumerate}
\section{Related Work}
\subsection{Multivariate Time Series Classification} 
Most MTSC methods can be grouped into three categories: distance-based, feature-based, and deep learning-based approaches. Here we only discuss deep learning-based strategies. 

Currently, two popular deep learning models, CNN and RNN, are widely used in MTS classification. These models often use an LSTM layer and stacked CNN layer to extract time-series features, and a softmax layer is then applied to predict the label \cite{zhang2020tapnet}. For example, MLSTM-FCN \cite{karim2019multivariate} utilizes an LSTM layer and a stacked CNN layer alongside squeeze-and-excitation blocks to obtain representations. TapNet \cite{zhang2020tapnet} also constructs an LSTM layer and a stacked CNN layer, followed by an attentional prototype network.

Deep learning-based methods require less domain knowledge in time series data than traditional methods. However, the limitation of the above models is obvious: they assume that the time series variables have the same effect among each other. 
Consequently, they cannot explicitly model the pairwise dependencies among variables.
In this case, graphs are the most appropriate data structure for modeling MTS.

\subsection{Graph Neural Networks} 
\cite{scarselli2008graph} first proposed the idea and the concept of graph neural network (GNN), which broadened existing neural networks for handling the information addressed in graph areas. 
GNNs follow a local aggregation mechanism, where the embedding vector of a node is processed by recursively aggregating and transforming embedding vectors of its neighbor nodes \cite{xu2021graph,duan2021connecting}. 
Numerous GNN variations have been proposed and have accomplished cutting-edge results on both node and graph classification assignments \cite{wang2019heterogeneous}. 
For example, the Graph Convolutional Networks (GCN) \cite{DBLP:journals/corr/KipfW16} could be viewed as an approximation of spectral-domain convolution of the graph data. 
GraphSAGE \cite{DBLP:journals/corr/HamiltonYL17}, and FastGCN \cite{chen2018fastgcn} sample and aggregate of the neighborhood information while enabling training in batches yet forfeiting some time-proficiency.
Graph Attention Networks (GAT) \cite{velivckovic2017graph} 
designs a new way to gather neighbors through self-attention.
After this, Graph Isomorphism Network (GIN) \cite{Xu2018HowPA} and k-GNNs \cite{morris2019weisfeiler} are developed, presenting more perplexing and different types of aggregation. 

\subsection{Graph Pooling} 
\label{Graph Pooling}
Graph pooling strategies can be characterized into three classes: topology based, global, and hierarchical pooling.
Here we only discuss hierarchical pooling methods.
DiffPool \cite{ying2018hierarchical} trains two parallel GNNs to obtain node-level representations and cluster assignments.
gPool \cite{gao2019graph}, and SAGPool \cite{lee2019self} drop nodes from the input graph as opposed to bunch various nodes to frame a cluster in the pooled graph.
They devise a top-K node choice method to make an initiated sub-graph for the following layers.
Although they are more efficient than DiffPool, they do not gather nodes nor calculate soft edge weights.
This makes them unable to preserve node and edge information effectively. 
MemGNN \cite{Khasahmadi2020Memory-Based} likewise soft cluster assignments, and they utilize a clustering-friendly distribution to figure the attention scores among nodes and clusters.
However, they generate centroids (which are used to calculate soft assignments) without involving the input graphs, which does not make sense intuitively because the centroids for different input graphs should be different.
We propose Variational Pooling, using an ``encoder-decoder'' architecture to get centroids to address this limitation. It can keep the property of permutation invariance while making centroids input-related to graphs \cite{xu2020cosimgnn}.

\section{Framework}
\begin{figure*}
\centering
\includegraphics[width=1\linewidth]{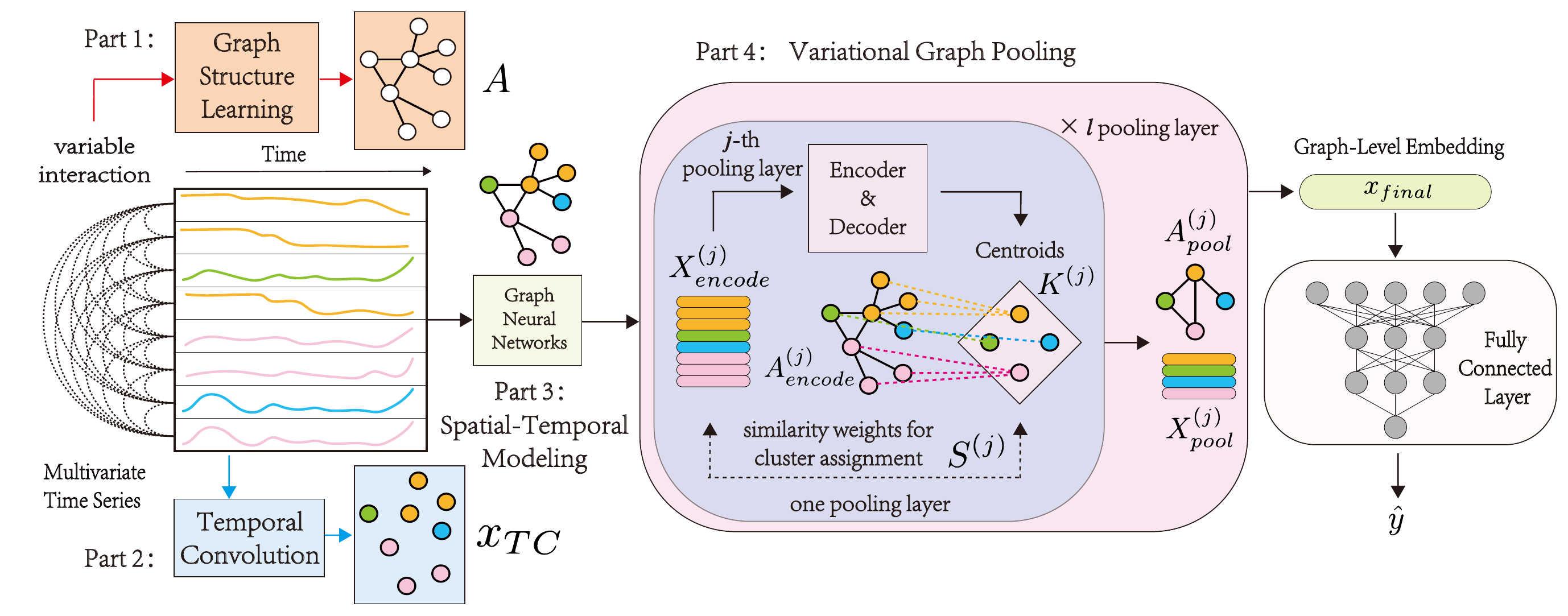}
\caption{The general architecture of MTPool. Adjacency matrix and feature matrix are constructed through graph structure learning and  temporal convolution respectively. Then GNNs aggregate and fuse spatial-temporal features. After several pooling layers, the input graphs are hierachically coarsened to one node to attain their graph-level representations. Finally, these final output graph-level embeddings can be used as features input to a differentiable classifier for MTSC tasks. MTPool is an end-to-end joint framework for graph structure learning, temporal convolution, graph representation learning and graph coarsening.}
\label{fig:1}
\end{figure*}

In this part, we present the proposed MTPool in detail. 
MTPool can be divided into four parts: Graph Structure Learning, Temporal Convolution, Spatial-temporal Modeling, and Variational Graph Pooling.
The schematic of MTPool is shown in Figure \ref{fig:1}. 

\subsection{Problem Formulation}
A multivariate time series (MTS) can be represented as a matrix 
$ \bm{X} = \{ \bm{x_1},  \bm{x_2},..., 
\bm{x_n}\}\in \mathbb{R}^{n \times T}$, 
in which $T$ is the length of the time series and $n$ is the number of MTS variables. 
Each MTS is associated with a class label $y$ from a predefined label set. 
Given a group of MTS 
$ \bm{\mathcal X} = \{\bm{X_1}, \bm{X_2},..., \bm{X_N} \}\in \mathbb{R}^{N \times n \times T}$,
where $N$ is number of MTS slices in the group,
and the corresponding labels 
$\mathcal Y = \{y_1, y_2,..., y_N\}\in \mathbb{R}^{N}$,
the research goal is to learn the mapping relationship between 
$\bm{\mathcal X}$
and 
$\mathcal Y$ 
based on the proposed model.

\subsection{Graph Structure Learning}
We propose a dynamic relation embedding strategy, which learns the adjacency matrix $\bm{A} \in \mathbb{R}^{n\times n}$ adaptively to model latent relations in MTS sample $\bm{X}$. The learned graph structure (dynamic adjacency matrix) $\bm{A}$ is defined as:
\begin{equation}
    \bm{A} = Embed_1(\bm{X}),
\end{equation}
where $Embed_1()$ represents the graph structure learning function.
For the input MTS $\bm{X} = \{\bm{x_1}, \bm{x_2}, ..., \bm{x_n}\} \in \mathbb{R}^{n\times T}$ 
we first calculate the similarity matrix $\bm{C}$ between sampled MTS variables:
\begin{equation}
    {C_{ij}}= \frac{exp\  \big(-\sigma(distance(\bm{x_i},\bm{x_j}))\big)}{\sum_{p = 1}^n exp \  \big(-\sigma(distance(\bm{x_i},\bm{x_p})\big)},
\end{equation}
where $\bm{x_i}$ and $\bm{x_j}$ denote the $i^{th}$ and $j^{th}$ MTS variables $(i,j = 1, 2, ..., n)$ and $distance$ denotes the distance metric such as Euclidean Distance, Absolute Value Distance, Dynamic Time Warping, etc.
Then the dynamic adjacency matrix $\bm{A}$ can be calculated as:
\begin{equation}
    \bm{A} = \sigma(\bm{C} \bm{W_{adj}}),
\end{equation}
where $\sigma$ is an activation function and $\bm{W_{adj}} \in \mathbb{R}^{n\times n}$ is learnable model parameters to dynamically generate adjacency matrix based on input features.
What's more, to improve training efficiency, reduce noise impact and make the model more robust, threshold $c_1$ is set to make the adjacency matrix sparse:

\begin{equation}
{A}_{ij}^{}=\left\{
\begin{aligned}
    {A}_{ij}^{}, &  & {A}_{ij}^{} >= c_1.\\
    0, &  & {A}_{ij}^{} < c_1.
\end{aligned}
\right.
\end{equation}
Finally, row normalization is applied to $\textbf{A}^{}$.

\subsection{Temporal Convolution} 
This stage aims to extract temporal features, which is associated with or change over time, and construct feature matrix $\bm{X_{TC}} \in \mathbb{R}^{n\times d}$, where $d$ is the obtained feature dimension. 
With temporal convolution, we can get each MTS's feature matrix as follows:
\begin{equation}
    \bm{X_{TC}} = Embed_2(\bm{X}),
\end{equation}
where $Embed_2()$ is the temporal convolution function.
When investigating time series, it is essential to consider its numerical value as well as its pattern over the long haul. 
Time series in the real world typically have numerous concurrent periodicities.
For instance, the number of inhabitants in a specific city shows a particular pattern each day. Yet, a significant example can be seen by noticing it on the size of multi-week or one month. 
In this manner, it is sensible and vital to separate the highlights of the time arrangement in units of numerous particular periods. 
To mimic the present circumstance, we utilize numerous CNN channels with various responsive fields, namely kernel sizes, to extract features at multiple time scales.

For the $i$-th CNN filters (the number of CNN filters is $q$ and $i = 1, 2, ..., q$),
given the input time series $\bm{X}$, the feature vector $\bm{f_i}$ are expressed as follows:
$\bm{f_i} = \sigma(\bm{W_i}*\bm{X}+\bm{b})$, where $*$ denotes the convolution operation, $\sigma$ is a nonlinear activation function, such as $RELU(x) = max(0,x)$, $\bm{W_i} \in \mathbb{R}^{1 \times ks}$ represents the $i$-th CNN kernel, $ks$ is the kernel size and $\bm{b}$ is the bias. 
And the final feature vector can be expressed as 
$\bm{X_{TC}} = \left(\mathop{\lVert}\limits_{i=1}^{|q|} \bm{f_i}\right)$, where $\mathop{\lVert}\limits_{i=1}^{|q|}$ means concatenate operation from the feature $\bm{f_1}$ to the feature $\bm{f_q}$.
In this way, temporal features under various periods are extracted, providing powerful reference for time series classification.

\subsection{Spatial-temporal Modeling}
In this stage, $m$ GNN layers $(G_{1},G_{2},...G_{m})$ are employed on the input graphs (denoted as $\bm{X_{TC}}, \bm{A}$) for spatial-temporal modeling.
GNN layers can fuse spatial dependencies and temporal patterns for embedding feature of nodes and transform the feature dimension of nodes to $d_{encode}$, just as shown in equation \eqref{abstracted equation for encoding}:
\begin{equation}
\label{abstracted equation for encoding}
\bm{Z} = G_m(G_{m-1}(...G_{1}(\bm{X_{TC}}, \bm{A})...)),
\end{equation}
where $\bm{Z}\in \mathbb{R}^{n \times d_{encode}}$, represents the learned node embedding from the spatial-temporal modeling.
$G_j\ (j=1,2,...,m)$ consists of a graph neural network layer (GNN) and a batch normalization layer.
GNN can be such as GCN \cite{NIPS2016_6081}, GAT \cite{velivckovic2017graph}, GIN \cite{Xu2018HowPA}, etc.
Inspired by k-GNNs \cite{morris2019weisfeiler} model, we use the following propagation mechanism in this paper to calculate the forward-pass update of a node denoted by $\bm{v_i}$:
\begin{equation}
\bm{z_i^{(j+1)}} = \sigma\Big( \bm{z_i^{(j)}}\bm{W_1^{(j)}}+ \sum_{r\in \bm{N(i)}}\bm{z_r^{(j)}}\bm{W_2^{(j)}} \Big),
\end{equation}
where 
$\bm{W_1^{(j)}}$ and $\bm{W_2^{(j)}}$ are parameter matrices of the $j$-th GNN layer, 
$\bm{z_i^{(j)}}$ is the hidden state of node $\bm{v_i}$ in the $j^{th}$ layer and $\bm{N(i)}$ denotes the neighbors of node $i$.
K-GNNs only perform information fusion between a specific node and its neighbors, ignoring the information of other non-neighbor nodes. 
This design highlights the relationship among variables, effectively avoiding the information redundancy brought by high dimensions. 

\subsection{Variational Graph Pooling}
\subsubsection{Overall Transformation}
In this stage, the encoded graph is hierarchically pool to a single node for generating its graph-level representation.
This successive process can be expressed as:
\begin{equation}
 \bm{x_{final}} = P_l(P_{l-1}(...P_{1}(\bm{Z,A})...)),
\end{equation}
where $P_j\ (j=1,2,...,l)$ represents one pooling layer, and 
after stacking $l$ pooling layers, we get its graph-level representation vector $\bm{x_{final}}$.
For the $j$-th pooling layer, $P_j()$ can pool each encoded graph, to a specific coarsened graph.
The overall transformation of $P_j()$ is shown in equation \ref{overall equations of pooling layer}:
\begin{equation}
\label{overall equations of pooling layer}
\begin{split}
  \bm{X_{pool}^{(j)}} &=\sigma\left( \bm{S^{(j)}} \bm{X_{encode}^{(j)}} \bm{W_{pool}^{(j)}} \right),   
\\
  \bm{A_{pool}^{(j)}} &=\sigma\left(  \bm{S^{(j)}} \bm{A_{encode}^{(j)}} \bm{(S^{(j)})^{T}} \right),
\end{split}
\end{equation}
where $\sigma$ is a non-linear activation function, $\bm{W_{pool}^{(j)}}\in \mathbb{R}^{d_{encode}^{(j)} \times d_{pool}^{(j)}}$ is a trainable parameter matrix standing for a linear transformation and $\bm{S^{(j)}}\in \mathbb{R}^{n_{pool}^{(j)}\times n_{encode}^{(j)}}$ is the assignment matrix representing a projection from the original nodes to pooled nodes (clusters).
The details of how to calculate $S^{(j)}$ are in the next subsection \ref{How to Compute Assignment Matrix}.
$\bm{X_{encode}^{(j)}} \in \mathbb{R}^{n_{encode}^{(j)} \times d_{encode}^{(j)}}$ and $\bm{A_{encode}^{(j)}} \in \mathbb{R}^{n_{encode}^{(j)} \times n_{encode}^{(j)}}$ represent the encoded feature and adjacency matrix of the input graph in the $j$-th pooling layer respectively; $\bm{X_{pool}^{(j)}} \in \mathbb{R}^{n_{pool}^{(j)} \times d_{pool}^{(j)}}$ and $\bm{A_{pool}^{(j)}} \in \mathbb{R}^{n_{pool}^{(j)} \times n_{pool}^{(j)}}$ represent the pooled feature and adjacency matrix of the pooled graph in the $j$-th pooling layer respectively.
In most cases, the pooled graph has less nodes than the input graph ($n_{pool}^{(j)}<n_{encode}^{(j)}$).
For the first pooling layer, we have: $\bm{X_{encode}^{(0)}} = \bm{Z}$ and $\bm{A_{encode}^{(0)}} = \bm{A}$.

\subsubsection{How to Compute Assignment Matrix}
\label{How to Compute Assignment Matrix}
In this part, we propose a new pooling method: Variational Pooling, to address the limitation of existing methods as we discussed in section \ref{Graph Pooling} and to calculate $\bm{S^{(j)}}$ in a more effective way.
Although there are many advanced graph pooling methods proposed in recent years, they all have some limitations.
For example, MemGNN \cite{Khasahmadi2020Memory-Based} utilizes a multi-head exhibit of memory keys and a convolution operator to sum the soft cluster assignments from various heads. 
MemGNN utilizes a clustering-friendly distribution to figure the attention scores among nodes and clusters and performs better than DiffPool in many tasks. Still, we notice that it generates memory heads, which stand for the new centroids in the space of pooled graphs, without the involvement of the input graphs.
However, centroids for different graphs should be different, and each input graph should have its corresponding centroids based on its topology structure and node features.
To address the limitations mentioned above, we first generate $h$ batches of centroids $\bm{K^{(j)}} = [\bm{K_1^{(j)}}, \bm{K_2^{(j)}}, ..., \bm{K_h^{(j)}} ]^T \in \mathbb{R}^{h\times n_{pool}^{(j)}\times d_{encode}^{(j)}}$ based on the input graph in $P_j()$ and then compute and aggregate the relationship between every batch of centroids and the encoded graph for assignment matrix $\bm{S^{(j)}}$.

\subsubsection{How to Generate Centroids}
The generation of centroids $\bm{K^{(j)}}$ should satisfy the following properties:
\begin{enumerate}[(1)]
\item \textbf{Permutation invariance.} 
The same graph can be addressed by various adjacency matrices by permuting the sequence for nodes.
The centroids of the same graph should be invariant to such changes.
\item \textbf{Input correlation.} The generation of centroids should be based on the input graph. If the input graph changes, the centroids should change accordingly to capture global features effectively.
\item \textbf{Dimensional adjustability.} The dimension of centroids matrix $\bm{K^{(j)}}$ should be adjusted according to the dimension of the input graph and the dimension of the output coarsened graph we need.
\end{enumerate}

Therefore, we propose an ``encoder-decoder'' architecture for computing centroids matrix $\bm{K^{(j)}}$. In general, the $encoder$ ensures the property of permutation invariance and makes the centroids adaptive to input graphs while the $decoder$ controls the dimension of the output coarsened graph:
\begin{equation}
\label{generation of memory heads}
\bm{K^{(j)}} = Decoder\left(Encoder\left( \bm{X_{encode}^{(j)}}\right)\right).
\end{equation}
As shown in equation \ref{generation of memory heads}, an $encoder$ is deployed over the input graph, transforming $\bm{X_{encode}^{(j)}}\in \mathbb{R}^{n_{encode}^{(j)} \times d_{encode}^{(j)}}$ to $\bm{X^{(j)}_{g}}\in \mathbb{R}^{1 \times d_{encode}^{(j)}}$.
Here $\bm{X_{g}^{(j)}}$ is the feature that incorporates the input information.
Then a $decoder$ is applied to map $\bm{X_{g}^{(j)}}$ to $\bm{K^{(j)}}\in \mathbb{R}^{(h\times n_{pool}^{(j)})\times d_{encode}^{(j)}}$, after which $\bm{K^{(j)}}$ is reshaped to $\mathbb{R}^{h\times n_{pool}^{(j)}\times d_{encode}^{(j)}}$. 
Refer to \cite{bai2019simgnn}, the expression of encoder and decoder we use in this paper is as follows:
\begin{equation}
\begin{split}
  Encoder:\bm{X_g^{(j)}} & = \sum_{i=1}^n \sigma \big(\bm{(u_i^{(j)})^T} \bm{x_{avg}^{(j)}} \bm{u_i^{(j)}}\big) \\
  & = \sum_{i=1}^n \sigma_1 \Big(\bm{(u_i^{(j)})^T} \sigma_2\big((\frac{1}{n} \sum_{j=1}^n \bm{u_j^{(j)}}) \bm{W_{avg}^{(j)}}\big) \bm{u_i^{(j)}}\Big), \\[1mm]
  Decoder &:MLP,
\end{split}
\end{equation}
where $\bm{u_i^{(j)}}$ is the embedding of node $i$ of the input graph in the $j$-th pooling layer, $\sigma_1$ is the $sigmoid$ function, $\sigma_2$ is the $tanh$ activation function and $\bm{W_{avg}^{(j)}} \in \mathbb{R}^{d_{encode}^{(j)}\times d_{encode}^{(j)}}$ is a learnable weight matrix.
For the encoder, we use the attention mechanism to guide the model to learn weights under specific tasks to generate graph-level representation $\bm{X_g^{(j)}}$ for centroids.
For decoder, here we use Multiple Layer Perceptron (MLP) to reshape matrix $\bm{K^{(j)}}$ to the size we need.

\subsubsection{Computing Assignment Matrix}
With centroids matrix $\bm{K^{(j)}}$, we can compute the relationship $\bm{S_p^{(j)}}\in \mathbb{R}^{n_{pool}^{(j)}\times n_{encode}^{(j)}}\ (p=1,2,...,h)$ between centroids $\bm{K_p^{(j)}}\in \mathbb{R}^{n_{pool}^{(j)}\times d_{encode}^{(j)}} \ (p=1,2,...,h)$ and $\bm{X_{encode}^{(j)}}\in \mathbb{R}^{n_{encode}^{(j)}\times d_{encode}^{(j)}}$ in $j$-th pooling layer.
We use cosine similarity to evaluate the relationship between input node embeddings and centroids, as described in equation \ref{dist calculation of pooling layers}, followed by a row normalization deployed in the resulting assignment matrix:
\begin{equation}
\label{dist calculation of pooling layers}
\begin{split}
\bm{(S_p^{(j)})'} &= \operatorname{cosine} \left(\bm{K_p^{(j)}}, \bm{X_{encode}^{(j)}}\right), \\[1mm]
\bm{S_p^{(j)}} &= \operatorname{normalize_{row}}\left(\bm{(S_p^{(j)})'}\right),
\end{split}
\end{equation}
where $\bm{(S_p^{(j)})'}\in \mathbb{R}^{n_{pool^{(j)}}\times n_{encode}^{(j)}}\ (p=1,2,...,h)$ is the assignment matrix before normalization.
We finally aggregate the information of $h$ relationship $\bm{S_p^{(j)}}$:
\begin{equation}
\label{conv on dist of pooling layers}
\bm{S^{(j)}} = \Gamma_{\phi}\left(\mathop{\lVert}\limits_{p=1}^{|h|} \bm{S_p^{(j)}}\right).
\end{equation}
In equation \ref{conv on dist of pooling layers}, we concatenate $\bm{S_p^{(j)}}\ (p=1,2,...,h)$ and perform a trainable weighted sum $\Gamma_{\phi}$ to the concatenated matrix, leading to the final assignment matrix $\bm{S^{(j)}}$.

\IncMargin{1em}
\begin{algorithm*} 
\label{Algorithm:1}
\SetKwData{Left}{left}
\SetKwData{This}{this}
\SetKwData{Up}{up} 
\SetKwFunction{Union}{Union}
\SetKwFunction{FindCompress}{FindCompress} \SetKwInOut{Input}{Input}
\SetKwInOut{Output}{Output}

\Input{The multivariate time series dataset of $N$ MTS slices,
$ \bm{\mathcal X} = \{\bm{X_1}, \bm{X_2},..., \bm{X_N}\}\in \mathbb{R}^{N \times n \times T}$;
} 
\Output{The corresponding predict labels, $\hat{\mathcal Y} = \{\hat y_{1}, \hat y_{2}, ..., \hat y_{N}\}\in \mathbb{R}^{N}$}
\BlankLine 

\emph{$\hat{\mathcal Y} = \{\}$}; 
\tcp{Record predict labels of MTS slices}

\For{$\bm{X_k}$ in $\{\bm{X_1}, \bm{X_2},..., \bm{X_N}\}, \bm{X_k} \in \mathbb{R}^{n \times T}$}{ 
   \For{$\bm{x_i}$ in $\bm{X_k}$ = $\{\bm{x_1}, \bm{x_2},..., \bm{x_n}\}, \bm{x_i} \in \mathbb{R}^{T}$}{
			\For{$\bm{x_j}$ in $\bm{X_k}$ = $\{\bm{x_1}, \bm{x_2},..., \bm{x_n}\}, \bm{x_j} \in \mathbb{R}^{T}$}{
				${C_{ij}}^{}= \frac{exp(-\sigma(distance(\bm{x_i},\bm{x_j})))}{\sum_{p = 0}^n exp(-\sigma(distance(\bm{x_i},\bm{x_p}))}$
				\tcp{Calculating the similarity matrix}
			}
		}
		$ \bm{A}^{} = \sigma(\bm{C}^{} \bm{W_{adj}})$;
		\tcp{Calculating the dynamic adjacency matrix}
		$ \bm{A}^{}$ = $Normalize_{row}(Threshold( \bm{A}^{}))$;
		\tcp{Graph Stucture Learning}
		$\bm{X^{}_{TC}} = Temporal\_Convolution(\bm{X_k}) \in \mathbb{R}^{n\times d}$;
		\tcp{Temporal Convolution}
		$\bm{Z} = G_m(G_{m-1}(...G_{1}(\bm{X^{}_{TC}}, \bm{A}^{})...))$;
		\tcp{Spatial-temporal Modeling}
		$\bm{X_{encode}^{(0)}} = \bm{Z}$, $\bm{A_{encode}^{(0)}} = \bm{A}$;
		\tcp{The input of the first pooling layer}
		\For{$P_j$ in $Pooling\_Layers$ = $\{P_{1}, P_{2},..., P_l\}$}{
		$\bm{K^{(j)}} = [\bm{K_1^{(j)}}, \bm{K_2^{(j)}}, ..., \bm{K_h^{(j)}} ]^T = Decoder\left(Encoder\left( \bm{X^{(j)}_{encode}}\right)\right)$\;
		$Encoder: \bm{X^{(j)}_{g}} = \sum_{i=1}^n \sigma_1 \Big(\bm{(u_i^{(j)})^T} \sigma_2\big((\frac{1}{n} \sum_{j=1}^n \bm{u_j^{(j)}}) \bm{W_{avg}^{(j)}}\big) \bm{u_i^{(j)}}\Big)$, $Decoder: MLP $\;
		\tcp{Generating Centroids}	
		$\bm{S^{(j)}} = \Gamma_{\phi}\left(\mathop{\lVert}\limits_{p=1}^{|h|} \bm{S^{(j)}_{p}}\right)$, $\bm{S^{(j)}_p} = Normalize_{row}\left(Cosine \left(\bm{K^{(j)}_{p}}, \bm{X^{(j)}_{encode}}\right)\right)$,
		$p=1, 2, ..., h$\;
		\tcp{Computing Assignment Matrix}
		$\bm{X_{pool}^{(j)}} =\sigma\left( \bm{S^{(j)}} \bm{X_{encode}^{(j)}} \bm{W_{pool}^{(j)}} \right)$\;
	  $\bm{A_{pool}^{(j)}} =\sigma\left(  \bm{S^{(j)}} \bm{A_{encode}^{(j)}} \bm{(S^{(j)})^{T}} \right)$\;
		\tcp{Variational Graph Pooling}
		$\bm{X_{encode}^{(j+1)}}=\bm{X_{pool}^{(j)}},\bm{A_{encode}^{(j+1)}}=\bm{A_{pool}^{(j)}}$;
	   \tcp{Get the input of the next pooling layer}
		}
	    $\bm{x_{final}}$ = $\bm{X_{pool}^{(j)}}$\;
		$\hat y_k$ = $MLP(\bm{x_{final}})$\;
  	$\hat{\mathcal Y} \leftarrow \hat y_k$\;
  	\tcp{Add $\hat y_k$ to $\hat{\mathcal Y}$}
} 
\Return{$\hat{\mathcal Y}$}
\caption{MTPool algorithm framework}
\end{algorithm*}
\DecMargin{1em} 

\subsection{Differentiable  Classifer}
What this stage do is to map the graph-level embedding vector $\bm{x_{final}}$ to a specific predicted class number $\hat y$. 
A standard $MLP$ is used to transform the dimension of the graph-level embedding to the number of classes.
Then we compare the predicted class number against the ground-truth label.
The loss function is as follows:
\begin{equation}
    \mathcal L = - \frac{1}{N} \sum_{i=1}^N \sum_{j=1}^M y_{i,j} log\hat y_{i,j},
\end{equation}
where $N$ is the set of training samples, $M$ denotes the number of classes, $y$ is the true label, and $\hat y$ is the value predicted by the model.
The whole algorithm framework is shown in algorithm 1.

\begin{table*}[t]
\centering
\caption{Summary of the 10 UEA datasets used in experiments.}
\renewcommand\arraystretch{1.5}
\scalebox{1}{
\begin{tabular}{|c|c|c|c|c|c|c|}
\hline
     & Name                  & Train Size & Test Size & Num Series & Series Length & Classes \\ \hline
AF   & AtrialFibrillation    & 15         & 15        & 2          & 640           & 3       \\
FM   & FingerMovements       & 316        & 100       & 28         & 50            & 2       \\
HMD  & HandMovementDirection & 160        & 74        & 10         & 400           & 4       \\
HB   & Heartbeat             & 204        & 205       & 61         & 405           & 2       \\
LIB  & Libras                & 180        & 180       & 2          & 45            & 15      \\
MI   & MotorImagery          & 278        & 100       & 64         & 3000          & 2       \\
NATO & NATOPS                & 180        & 180       & 24         & 51            & 6       \\
PD   & PenDigits             & 7494       & 3498      & 2          & 8             & 10      \\
SRS2 & SelfRegulationSCP2    & 200        & 180       & 7          & 1152          & 2       \\
SWJ  & StandWalkJump         & 12         & 15        & 4          & 2500          & 3       \\ \hline
\end{tabular}}
\label{sec:datasets}
\end{table*}

\section{Experiments}
In this section, we lead comprehensive analyses on ten benchmark datasets for MTS classification and look at the evaluation results of our model (MTPool) with other baselines.

\subsection{Experiment Settings}
\subsubsection{Datasets}
We use ten publicly available benchmark datasets from the UEA MTS classification archive.\footnote{Datasets are available at http://timeseriesclassification.com. We exclude data with extremely long length, unequal length, high dimension size, and in-balance split.}
The main characteristics of each dataset are summarized in Table \ref{sec:datasets}. 
The train and test sizes represent the number of MTS slices in the train and test datasets, respectively.
Num series refers to the number of variables in each MTS slice; 
the series length refers to the length or the feature dimension of each variable in each MTS slice; 
class refers to the number of types of MTS slices.

\subsubsection{Metrics}
Like other MTS classification methods \cite{zhang2020tapnet}, we use classification accuracy as an evaluation metric:
\begin{equation}
Accuracy=\frac{TP+TN}{TP+TN+FP+FN},
\end{equation}
where $TP$, $TN$, $FP$ and $FN$ respectively stands for true positive, true negative, false positive and false negative.

\subsubsection{Methods for Comparison}
We use the following implementations of the MTS classifiers, including the common
distance-based classifiers, the latest bag-of-patterns model and the deep learning models: 
\begin{enumerate}[(1)]
    \item \textbf{ED, DTW$_{I}$, DTW$_{D}$, - with and without normalization (norm)} \cite{bagnall2018uea} are the common distance-based models. 1-Nearest Neighbour with distance functions include Euclidean (ED); dimension-independent dynamic time warping (DTW$_{I}$); and dimension-dependent dynamic time warping (DTW$_{D}$).
    \item \textbf{WEASEL+MUSE} \cite{schafer2017multivariate} stands for the Word ExtrAction for time SEries cLassification (WEASEL) with a Multivariate Unsupervised Symbols and dErivatives (MUSE). It is the most effective bag-of-patterns algorithm for MTSC.
    \item \textbf{HIVE-COTE} \cite{bagnall2020tale}
    is a heterogeneous meta ensemble for time series classification. This approach is a good baseline for assessing bespoke Multivariate Time Series Classification.
    \item \textbf{MLSTM-FCN} \cite{karim2019multivariate} is a famous deep-learning framework for MTSC. It utilizes an LSTM layer and a stacked CNN layer alongside squeeze-and-excitation blocks to obtain representations.
    \item \textbf{TapNet} \cite{zhang2020tapnet} draws on the strengths of both traditional and deep learning approaches. It also constructs a LSTM layer and a stacked CNN layer, followed by an attentional prototype network.
    \item \textbf{MTPool} is the MTPool framework with our proposed Variational Pooling and dynamic adjacency matrix.
\end{enumerate}

\subsubsection{Training Details}
All the networks are implemented with Pytorch 1.4.0 in python 3.6.2 and trained with 10000 epochs 
(computing infrastructure: Ubuntu 18.04 operating system, GPU NVIDIA GeForce RTX 2080 Ti with 8 Gb GRAM and 32 Gb of RAM). 
We use \{3, 5, 7\} as the convolutional kernel size and ten as the channel number.
The threshold to make the adjacency matrix sparse is chosen from \{0.05, 0.1, 0.2\} and the output dimension of the GNNs layer is 128.
For pooling layers, heads of centroids are chosen from \{1,2,4\}, the reduction factor is chosen from \{2, 3, 6\} and the number of nodes in the final pooling layer is 1.
The initial learning rate is $10^{-4}$, and the categorical cross-entropy loss and the Adam optimization are used to optimize the parameters of our models.

\begin{table*}[t]
\centering
\caption{Accuracy of the 11 algorithms on the default training/test datasets of 10 selected UEA MTSC archives. The best performance is bolded.}
\renewcommand\arraystretch{1.5}
\scalebox{0.85}{
\begin{tabular}{|c|cccccccccc|c|c|}
\hline
Methods / Datasets     & AF             & FM
& HMD            & HB             & LIB          & MI            & NATO           & PD             & SRS2          & SWJ   	& Avg. Rank	& Wins/Ties            \\ \hline

ED           & 0.267          & 0.519         & 0.279          & 0.620           & 0.833        & 0.510          & 0.850  & 0.973          & 0.483         & 0.333                  & 7.50	& 0 \\ \hline

DTW$_I$        & 0.267          & 0.513         & 0.297          & 0.659          & 0.894        & 0.390          & 0.850           & 0.939          & 0.533         & 0.200     	& 7.70	& 0      \\ \hline

DTW$_D$        & 0.267          & 0.529         & 0.231          & 0.717          & 0.872        & 0.500           & 0.883          & 0.977          & 0.539         & 0.200      	& 6.50	& 0              \\ \hline

ED (norm)    & 0.200            & 0.510          & 0.278          & 0.619          & 0.833        & 0.510          & 0.850           & 0.973          & 0.483         & 0.333    	& 8.25	& 0      \\ \hline

DTW$_I$ (norm) & 0.267          & 0.520          & 0.297          & 0.658          & 0.894        & 0.390          & 0.850           & 0.939          & 0.533         & 0.200       	& 7.60	& 0      \\ \hline

DTW$_D$ (norm) & 0.267          & 0.530          & 0.231          & 0.717          & 0.870         & 0.500           & 0.883          & 0.977          & 0.539         & 0.200       	& 6.50	& 0   \\ \hline

WEASEL+MUSE  & 0.400            & 0.550          & 0.365          & 0.727          & 0.894        & 0.500           & 0.870           & 0.948 & 0.460          & 0.267                 & 5.65	& 0\\ \hline

HIVE-COTE    & 0.133          & 0.550          & 0.446          & 0.722          & \textbf{0.900} & 0.610          & 0.889          & 0.934          & 0.461         & 0.333          & 5.40	& 1   \\ \hline

MLSTM-FCN    & 0.333          & 0.580 & \textbf{0.527} & 0.663          & 0.850         & 0.510          & 0.900            & 0.978 & 0.472         & 0.400      	&4.40	& 1       \\ \hline

TapNet       & 0.200            & 0.470          & 0.338          & \textbf{0.751} & 0.878        & 0.590          & 0.939          & 0.980           & 0.550          & 0.133     	&5.25	& 1     \\ \hline

MTPool     & \textbf{0.533}          & \textbf{0.620}          & 0.486          & 0.742          & \textbf{0.900}        & \textbf{0.630} & \textbf{0.944}          & \textbf{0.983} & \textbf{0.600}  & \textbf{0.667}     & \textbf{1.25}	& \textbf{8}     \\ \hline

\end{tabular}
}
\label{sec:table}
\end{table*}

\subsection{Main Results}

Tabel \ref{sec:table} shows the accuracy results on the selected 10 UEA datasets of MTPool and other MTS classifiers.
"Avg. Rank" in the table means the average ranking of different models under all datasets, and "Wins/Ties" indicates the number of datasets for which this model achieves the best performance.
For some approaches that ran out of memory, we refer to \cite{ruiz2020benchmarking} for the corresponding results.
The best performance in each dataset is bolded.

First of all, we can observe that MTPool wins on eight datasets, better than several other advanced MTS classification methods (such as TapNet or MLSTM-FCN wins on one dataset).
The "Avg. Rank" indicates the superiority of MTPool over the existing state-of-the-art models (the second-highest ranked model, MLSTM-FCN, has an average ranking of 4.40).
We can also find that methods based on deep learning proposed in recent years, such as TapNet and MLSTM-FCN, have better performance than methods based on nearest neighbors, proving the effectiveness of deep learning in extracting MTS features.

More importantly, with the different datasets, the number of variables and lengths span an extensive range. 
For instance, the variables number ranges from 2 to 64, and the minimum length of MTS is eight, and the maximum is 300.
In the case of variable dataset parameters, our framework maintains considerable competitiveness, proving our model's robustness.
More specifically, for a large number of variables, MTPool can handle this by increasing the pooling layers and making the graph pooling process more hierarchical, thus performing well.
For instance, MTPool achieved the best and second-best accuracy on the MotorImagery (64 variables) and Heartbeat (61 variables).
This means that when the number of variables is relatively large, the hierarchical pooling framework we provide can better capture the graph's structure and generate time series embeddings with more robust characterization capabilities.
It is worth noting that when the number of variables is relatively small, by reducing the number of pooling layers, MTPool can also achieve good accuracy: MTPool obtains the best accuracy on PenDigits (2 variables), SelfRegulationSCP2 (7 variables), and StandWalkJump (4 variables) datasets.
A more detailed analysis of MTPool is in the following subsection.

\begin{figure}[t]
\centering
\subfigure[Ablation study of pooling methods]{
\begin{minipage}[t]{0.48\linewidth}
\centering
\includegraphics[width=1\linewidth]{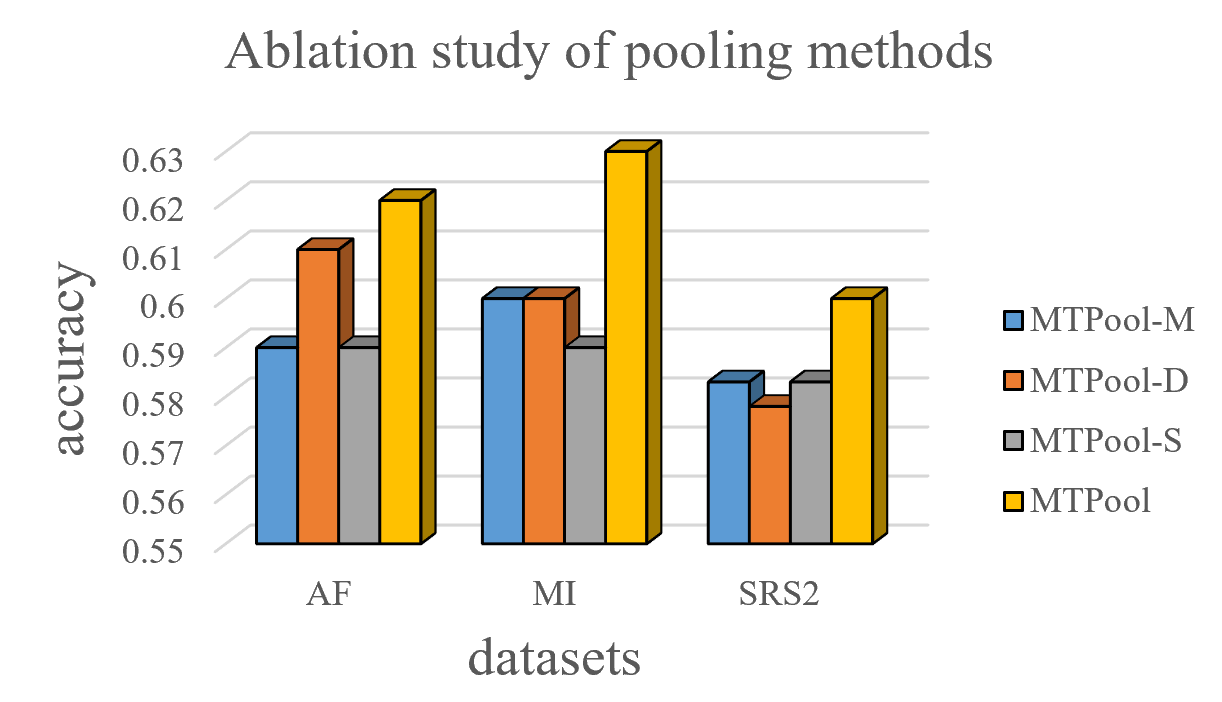}
\label{ablation study: pool}
\end{minipage}
}
\subfigure[Ablation study of adjacency matrix]{
\begin{minipage}[t]{0.48\linewidth}
\centering
\includegraphics[width=1\linewidth]{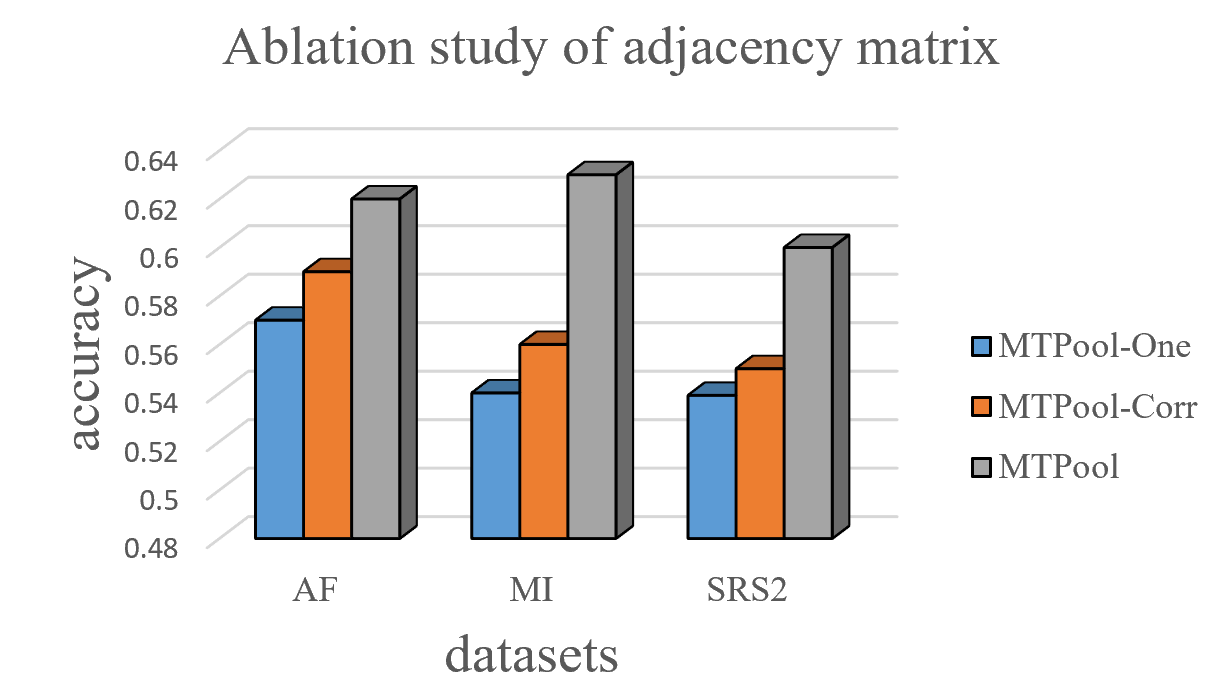}
\label{ablation study: adj}
\end{minipage}
}
\caption{Ablation study of pooling methods and adjacency matrix. AF, MI and SRS2 datasets are used and different colors represent different methods.}
\label{ablation study}
\end{figure}

\subsection{Ablation Study}
We conducted an ablation study to validate the effectiveness of key components that contribute to the improved outcomes of MTPool. 
First, we substitute our Variational Pooling layers with some other advanced pooling methods in the MTPool framework.
Then we use a static adjacency matrix for the initial graph structure instead of a dynamic adjacency matrix.
The detailed setting of each variant model is as follows.

\begin{itemize}
    \item \textbf{MTPool-M} is the MTPool framework with MemPool \cite{Khasahmadi2020Memory-Based}, which generate clustering centroids without involving the input graphs.
    \item \textbf{MTPool-D} is the MTPool framework with DiffPool \cite{ying2018hierarchical}, which trains two parallel GNNs to get node-level embeddings and cluster assignments.
    \item \textbf{MTPool-S} is the MTPool framework with SAGPool \cite{lee2019self}, which drops nodes from the input to pool the graph.
    \item \textbf{MTPool-One} is the MTPool framework with Variational Pooling and all-one adjacency matrix.
    \item \textbf{MTPool-Corr} is the MTPool framework with Variational Pooling and correlation coefficient adjacency matrix.
    \item \textbf{MTPool} is the MTPool framework with our proposed Variational Pooling and dynamic adjacency matrix.
\end{itemize}

Figure  \ref{ablation study} shows the comparison results.
The important conclusions of these results are as follows:
\begin{enumerate}[(1)]
    \item  Different hierarchical graph pooling methods can be incorporated in our MTPool framework and can achieve comparable or better performance than state-of-the-art MTSC methods.
    \item Different adjacency matrices can be used in our MTPool framework. However, the all-one matrix's performance is slightly worse than the correlation coefficient matrix, and our dynamic matrix achieves the best.
    \item Even if the three cutting-edge pooling methods (MemPool, DiffPool, and SAGPool) and two adjacency matrices (all-one and corr) have their own merits in the three selected datasets, our proposed Variational Pooling can achieve the best performance in all cases. This proves the effectiveness of our well-designed pooling method and the dynamic adjacency matrix.
\end{enumerate}


\begin{figure}[t]
\centering
\includegraphics[width=0.7\linewidth]{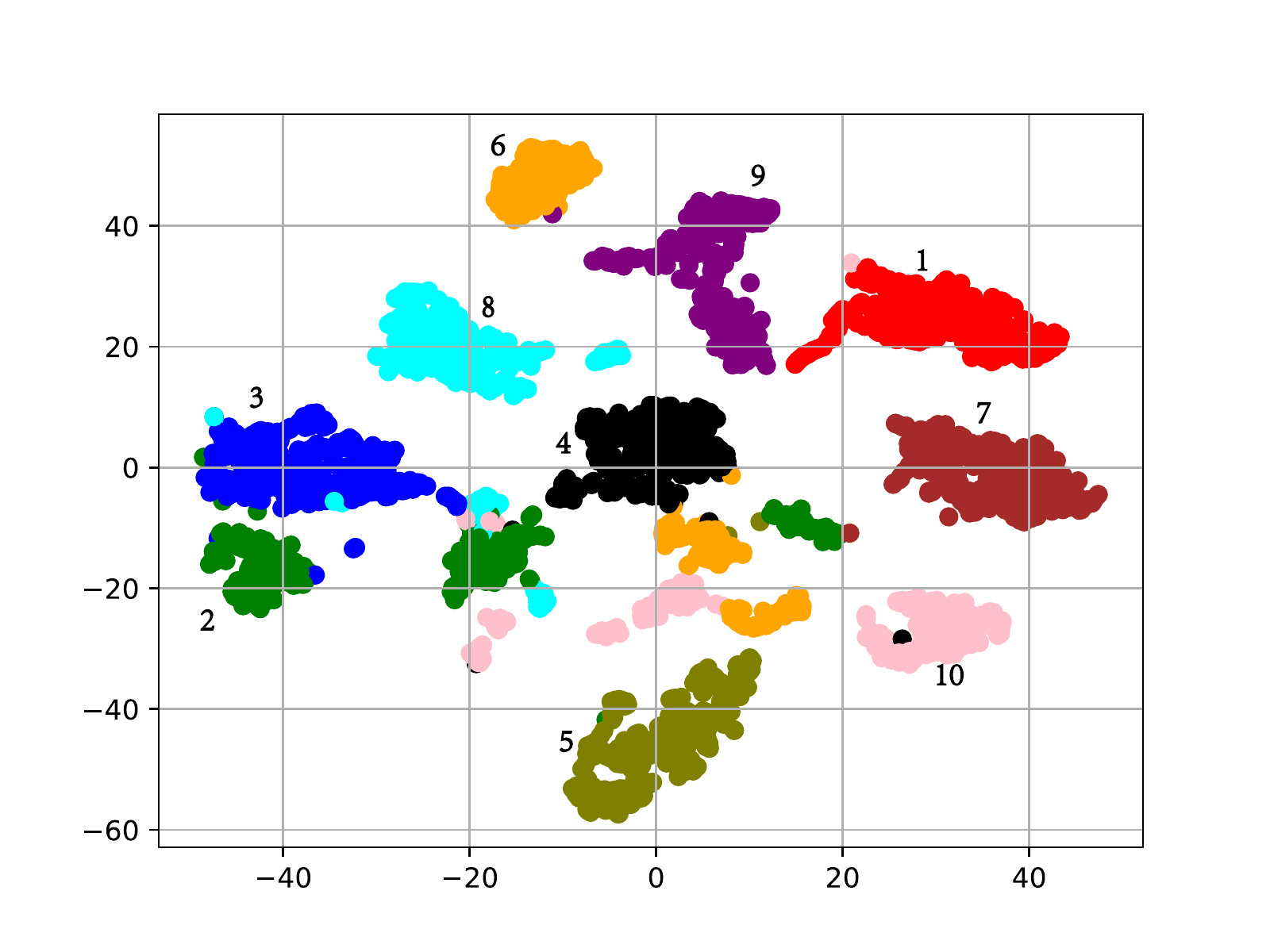}
\caption{Class Prototype Inspection: visualize the 128-dimension multivariate time series embeddings learned for the PenDigits dataset in a two-dimensional image by t-SNE.}
\label{fig:3}
\end{figure}

\subsection{Inspection of Class Prototype}
This section visualizes the class prototype and its corresponding time series embedding to prove our well-trained time series embedding effectiveness.
We use the t-SNE algorithm \cite{maaten2008visualizing} to visualize the 128-dimensional time series embedded in the form of two-dimensional images. We use different colors to distinguish different categories.
Figure 3 shows the embeddings learned for the PenDigits dataset, containing 3498 test samples in 10 different types.
The following conclusions can be drawn from the results: 
\begin{enumerate}[(1)]
    \item  The distance between data samples from different categories is much greater than the distance between data samples from the same type. Therefore, we can easily use the learned multivariate time series to embed the time sequence classification.
    \item Low-dimensional time series embedding provides us with a more interpretable perspective to understand classifiers' problems. For example, we see that categories two, six, and ten are not divided into complete pieces. Instead, these three categories are all divided into several sub-parts. Thus, it helps identify problems with the classifier and take further measures, such as adding more training samples in these three classes.
\end{enumerate}

\begin{figure}[t]
\centering
\includegraphics[width=1\linewidth]{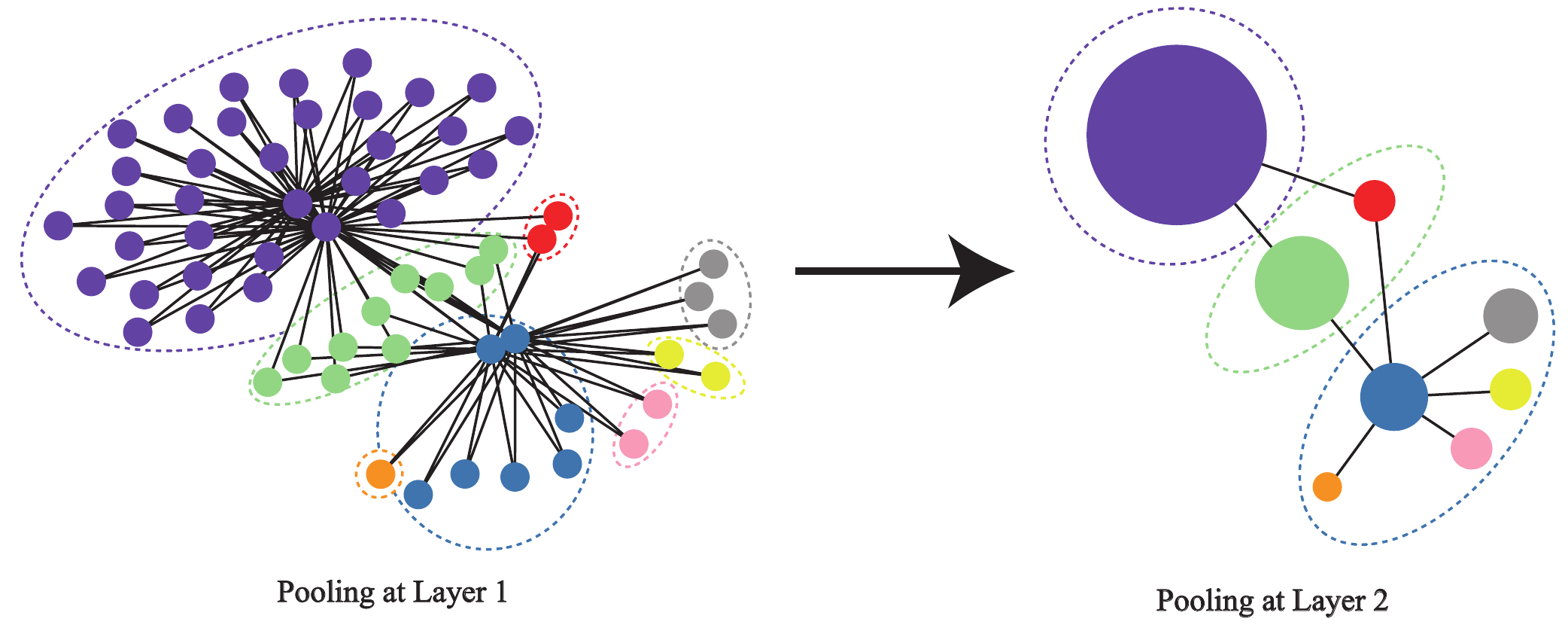}
\caption{Visualization of the graph pooling process in MTPool, using example graphs from Heartbeat dataset, which has 61 MTS variables, so the original graph has 61 nodes. Nodes in the second layer correspond to clusters in the first layer. We use the same color to represent nodes of the same cluster and use dotted lines to indicate different clusters.}
\label{fig:4}
\end{figure}

\subsection{Case Study: Visualizations}
In this section, we visualize the graph pooling process by using the Heartbeat dataset.
Figure \ref{fig:4} shows a visualization of node assignments in the first and second layers on a graph constructed from the Heartbeat dataset.
We use the same color to represent nodes of the same cluster.
The cluster membership of each node is determined by the argmax of its cluster assignment probabilities.
We also observed that even if the final goal is to get the graph-level embedding and get the class to which MTS belongs, MTPool can still capture the hierarchical graph structure, which helps us further reveal the dependencies among different variables in MTS.
It is worth mentioning that the assignment matrix may not assign nodes to specific clusters.
The column corresponding to the unused cluster has a lower value for all nodes.
For example, in this case, the expected number of clusters we set in the first layer is 15 (greater than 8), but in fact, we get 8 clusters.
This reminds us that even if we define the expected cluster number in advance, MTPool will automatically perform clustering to get the best coarser results suitable for this graph.
Such characteristics can be adjusted for different inputs, thus having a strong generalization ability.

\section{Conclusion}
In this paper, we propose the first graph pooling-based framework, MTPool, for MTS classification.
It can explicitly model the pairwise dependencies among MTS variables and attain global embedding with strong interpretability and expressiveness.
Experimental results demonstrate our model achieves state-of-the-art performance in the existing models.

Future research is promising to explore and design more powerful hierarchical graph pooling approaches that can be incorporated into our MTPool framework to attain a more expressive and interpretable global representation for MTS.

\section{Acknowledgments}
This work is supported in part by the National Key Research and Development Program of China (No.2019YFB2102600), the National Natural Science Foundation of China (No.62002035), the Natural Science Foundation of Chongqing(No.cstc2020jcyj-bshX0034).

\bibliographystyle{cas-model2-names}

\bibliography{cas-sc-template}

\end{document}